\documentclass[runningheads]{llncs}
\usepackage{graphicx}
\usepackage{listings}
\usepackage{listings}

\usepackage{xcolor}
\usepackage{amssymb}  
\usepackage{mathrsfs} 
\usepackage{multirow}

\usepackage{amsmath}

\usepackage{graphicx}
\usepackage[colorlinks=true, allcolors=blue]{hyperref}

\definecolor{codegreen}{rgb}{0,0.6,0}
\definecolor{codegray}{rgb}{0.5,0.5,0.5}
\definecolor{codepurple}{rgb}{0.58,0,0.82}
\definecolor{backcolour}{rgb}{0.95,0.95,0.92}

\lstdefinestyle{mystyle}{
  backgroundcolor=\color{backcolour}, commentstyle=\color{codegreen},
  keywordstyle=\color{black},
  numberstyle=\tiny\color{codegray},
  stringstyle=\color{codepurple},
  basicstyle=\ttfamily\scriptsize,
  breakatwhitespace=false,         
  breaklines=true,                 
  captionpos=b,                    
  keepspaces=true,                 
  numbers=left,                    
  numbersep=5pt,                  
  showspaces=false,                
  showstringspaces=false,
  showtabs=false,                  
  tabsize=2
}
\usepackage{amssymb}
\usepackage{pifont}

\begin{document}
\title{GPT4MIA: Utilizing Generative Pre-trained Transformer (GPT-3) as A Plug-and-Play Transductive Model for Medical Image Analysis}
%
\author{Yizhe Zhang\inst{1} \and Danny Z. Chen \inst{2}}
%
%
\institute{School of Computer
Science and Engineering, Nanjing University of Science and Technology, Nanjing, Jiangsu 210094, China \\\email{yizhe.zhang.cs@gmail.com} \and Department of Computer Science and Engineering, University of Notre Dame, Notre Dame, IN 46556, USA \\\email{dchen@nd.edu}}
%
\titlerunning{Utilizing GPT-3 for Medical Image Analysis}
\maketitle              
\begin{abstract}



In this paper, we propose a novel approach (called GPT4MIA) that utilizes Generative Pre-trained Transformer (GPT) as a plug-and-play transductive inference tool for medical image analysis (MIA). We provide theoretical analysis on why a large pre-trained language model such as GPT-3 can be used as a plug-and-play transductive inference model for MIA. At the methodological level, we develop several technical treatments to improve the efficiency and effectiveness of GPT4MIA, including better prompt structure design, sample selection, and prompt ordering of representative samples/features. We present two concrete use cases (with workflow) of GPT4MIA: (1) detecting prediction errors and (2) improving prediction accuracy, working in conjecture with well-established vision-based models for image classification (e.g., ResNet). Experiments validate that our proposed method is effective for these two tasks. We further discuss the opportunities and challenges in utilizing Transformer-based large language models for broader MIA applications.


 

\end{abstract}
\section{Introduction}



%



Modern large language models (LLMs) are built based on the Transformer architecture and are trained to produce a sequence of text output given a sequence of text input such that the output is expected to be semantically \textbf{coherent} to the input. For example, for a text completion task, the input text is a sequence of text from a text resource, and the model is trained to produce the next character, word, or sentence of the input text. Open AI's GPT-3 has 175 billion parameters, and was trained on hundreds of billions of words. Brown et al.~\cite{brown2020language} showed that GPT-3 is capable of few-shot learning: Given a few examples/demonstrations to GPT-3, it can generalize considerably well to new samples with similar characteristics. The input and output coherency and the strong generalization capability indicates that pre-trained LLMs such as GPT-3 are potentially capable as general tools for \textbf{transductive inference} tasks with limited data. 





\begin{figure}[t]
\centering
\includegraphics[width=0.8\textwidth]{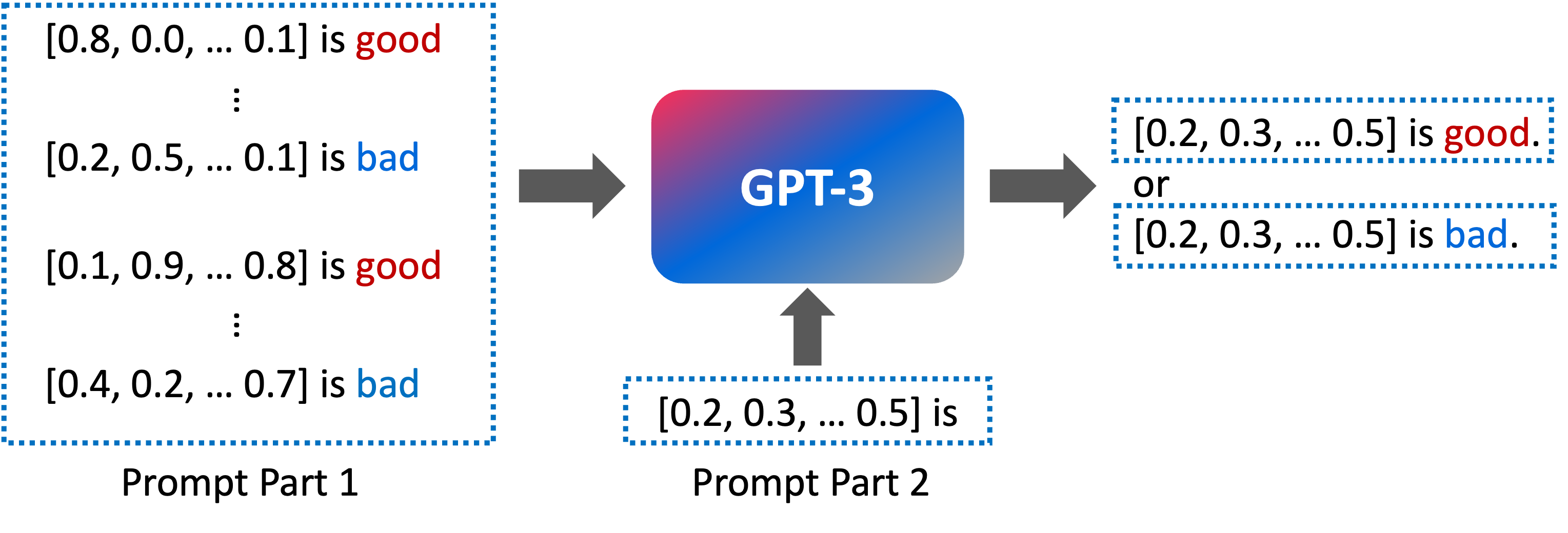}
\caption{Illustrating our high-level idea: Using GPT-3 for transductive inference on a binary classification task. Feature texts in Prompt Part 1 are from a set of samples with known labels. Prompt Part 2 contains feature text of a test sample.} \label{fig:overview}
\end{figure}

The notion of transductive inference was first introduced by Vapnik \cite{DBLP:books/daglib/0097035}. Given training samples (with labels) and test samples, transductive inference predicts the labels of the test samples using either a parametric model (e.g., a transductive support vector machine (SVM)~\cite{joachims1999transductive}) or a non-parametric model (e.g., a nearest neighbor based classifier~\cite{hang2022under}). Different from inductive inference, transductive inference does not aim to induce a prediction function from known samples; instead, its goal is to obtain the labels of test samples via propagating the information from known samples (e.g., training samples). 

In this paper, we propose a novel approach, called GPT4MIA, which utilizes GPT-3 as a plug-and-play transductive model to improve medical image analysis (MIA). For an MIA task (e.g., medical image classification), we give information of \textbf{known} samples as part of GPT-3's input and ask GPT-3 to infer a new sample's label (see Fig.~\ref{fig:overview}). We expect GPT-3 to infer a test sample's label by using transductive information from the known samples on the test sample. We give theoretical analysis on why this approach is feasible by drawing connections between attention mechanism and nearest neighbor inference. To make this approach more efficient and effective, we optimize the prompt construction, aiming to choose the most representative samples/features and order them in the prompt based on their importance. We present two practical use cases of utilizing our proposed method in medical image classification. We then validate the effectiveness of our method on medical image classification benchmarks. 

Our method utilizes a generative pre-trained Transformer for performing transduction from known medical image samples (e.g., training samples) to new test samples. The GPT-3 used in this work has billions of parameters. However, these parameters were pre-trained with language (text) data, and are not being updated during the transduction process for medical image classification. To our best knowledge, this is the first study to utilize a large pre-trained Transformer-based language model for performing transductive inference for image classification tasks (computer vision tasks), which are out of the data domain of the pre-training (language) domain. Our contributions are summarized as follows.

(1) We propose to utilize a large pre-trained language model (e.g., GPT-3) as a plug-and-play transductive inference method for improving MIA. We show that GPT-3 can serve as a general tool for performing transduction with an appropriate setup. Our approach is novel and flexible, suggesting a new direction of research for improving medical AI's accuracy and reliability. 

(2) We develop techniques to improve the efficiency, effectiveness, and usability of our proposed GPT4MIA. Two use cases are proposed for GPT4MIA, and strong empirical results validate that GPT4MIA outperforms conventional and state-of-the-art methods in both inductive and transductive method categories.

(3) Our work offers a new way of utilizing a small set of additional labeled data in medical AI: Given a trained deep learning (DL) model and a small set of labeled data (e.g., a validation set), utilizing GPT-3 as a transductive inference method in conjunction with a DL model can achieve better prediction reliability (use case \#1) and higher prediction accuracy (use case \#2).







\section{Approach}

In this section, we first provide theoretical analysis on the connection between the attention mechanism and transductive inference mechanism. Then we show details on how to design prompts for using GPT-3 as a transductive inference method. Finally, we present two use cases with workflow to demonstrate how to use GPT-3 as a plug-and-play transductive inference method for MIA.




\subsection{Theoretical Analyses}
A fundamental component of GPT-3 is the scaled dot-product attention. Typically, three pieces of input are fed to an attention layer: queries $Q$, keys $K$, and values $V$. The scaled dot-product attention can be described as:
\begin{equation}
Attention (Q,K,V) = softmax (\frac{QK^T}{s})V,
\label{eq:1}
\end{equation}
where $s$ is a scaling factor. Below, we show that a special case of Eq.~(\ref{eq:1}) can be viewed as a nearest neighbor (NN) classifier under a cosine distance metric system\footnote{Nearest neighbor classifiers are a typical transductive method for prediction problems.}.

\textbf{Setup 1:} Suppose the key component $K$ contains features of a set of $m$ known samples, and each feature is of a unit length. The value component $V$ contains these $m$ samples' corresponding labels, and each label is a one-hot vector. The query component $Q$ contains a feature vector (of a unit length), which represents a new test sample whose label is yet to be determined. 

\textbf{Proposition 1:} When the scaling factor $s$ is approaching 0 (e.g., $s$ is a very small positive number), the attention function in Eq.~(\ref{eq:1}) is approaching an NN classifier in the cosine distance metric system. 

The above is not difficult to show. $QK^T$ computes the pair-wise similarities between the test sample's feature and the features in the keys $K$. A small $s$ would 
enlarge the numerical gap between similar pairs and dissimilar pairs. This then leads to a one-hot-like result after applying the $softmax$ operation. The one-hot-like result is then multiplied with the values $V$, which chooses the label of a known sample that is the most similar to the test sample.

Generative Pre-trained Transformer uses a special type of attention called ``self-attention'', where the $K$, $V$, and $Q$ components are all the same. We will show that in a slightly different setup from Setup 1, the self-attention mechanism can also serve a role as an NN classifier for inferring a new sample's label given known samples' information.

\textbf{Setup 2:} For each known sample, we concatenate its feature vector with the corresponding label vector to form a feature-label vector. We repeat this process for every known sample, and put all the obtained feature-label vectors into $K$ (row by row). In addition, we construct the test sample's feature-label vector by concatenating its feature vector with a label vector containing all zeros. We put this feature-label vector into $K$ as well. Since we are considering the self-attention mechanism, $V$ and $Q$ are constructed in the same way as for $K$. 

\textbf{Proposition 2:} Under Setup 2, self-attention (i.e., Eq.~(\ref{eq:1}) with $K=V=Q$) generates the same label vector as one that is generated from the attention in Setup 1 for the test sample. With $s$ approaching a small value, self-attention can serve as an NN classifier for inferring the test sample's label.

Since the label vector for the test sample has all zeros at the input, the similarity measures between the test sample and known samples are influenced only by their features. This leads the inference process for the label of the test sample to be essentially the same as shown in Proposition 1. Transformer architecture used in modern large language models, including GPT-3, consists of multiple layers of self-attentions. Below we give more results on stacking self-attentions.




 \textbf{Proposition 3:} Under Setup 2, a single layer of self-attention (Eq.~(\ref{eq:1}) with $K=V=Q$) performs one iteration of clustering on feature-label vectors (including the known samples and test sample). $L$ layers of self-attention perform $L$ iterations of clustering. There exists a number $L^*$ for the number of layers of self-attention for which the clustering process converges. 

Guided by the above theoretical analysis, below we proceed to design the prompt (input) of GPT-3 for transductive inference. We use Setup 2 to guide the prompt construction since GPT-3 uses self-attention: The features and labels of the known samples and the feature of the test sample are put together to feed to GPT-3. According to Proposition 3, stacking self-attentions is functionally more advanced than a nearest neighbor-based classifier. GPT-3 uses not only stacking self-attentions but also numerous pre-trained parameters to augment these attentions. Hence, we expect GPT-3 to be more robust than the conventional methods (e.g., KNN) for transductive inference.



\subsection{Prompt Construction}\label{sec:promptdesign}

A set of $m$ known samples is provided with their features $F=\{f_1, f_2, \dots, f_m\}$ and corresponding labels $Y=\{y_1, y_2, \dots, y_m\}$.  A feature vector $f_{test}$ of a test sample is given. The task in this section is to construct a prompt text representation that contains information from $F$, $Y$, and $f_{test}$, which is fed to GPT-3 for inferring the label of the test sample. 

\textbf{Selecting and Ordering Known Samples.} As a language model, the original goal of training GPT-3 was to train the model to generate output that is semantically coherent with its input. The data used for training GPT-3 implicitly imposed a prior: The later a text appears in the input prompt (the closer the text to the output text), the larger impact it would impose on the output generation process. Hence, it is essential to put the more representative feature-label texts near the end of the prompt for inferring the test sample's label.


We compute pair-wise similarities between the features in the set $F$ of the known samples and obtain an affinity matrix $\textrm{S}$, in which each entry $\textrm{S}_{i,j}$ describes the similarity between samples $i$ and $j$ and is computed as $sim(f_i, f_j)$. A cosine similarity function is the default choice for $sim(.,.)$.

For a feature vector $f_i\in F$, we define a simple measure of how well $f_i$ represents the other known samples:
$\textrm{rep}_i= {\sum_{j=1}^{m}\textrm{S}_{i,j}}$. To select the top $k$ representative samples, one can compute $\textrm{rep}_i$ for each $i=1,2, \dots, m$, and choose the largest $k$ representative samples:
$\textbf{index}=argsort(\textrm{rep}_1, \dots, \textrm{rep}_m, ``descend")$, and
$\textbf{index}$ is represented as $\textbf{index}[1,2,\dots,k]$. The order of the samples in the prompt for GPT-3 should be in the reverse order of that in the $\textbf{index}$ list, where the most representative sample ($f_{index[1]}$) should be put at the end of the prompt in order to give more influence on the output generation. When dealing with imbalanced classification problems, we perform the above process for the samples in each class, and join them in an interleaved fashion.


\textbf{Converting Features and Labels to Feature-label Texts.}
For all the feature vectors $f_{i}$ where $i$ is in the $\textbf{index}$ list computed above, we convert these features to texts in an array-like format. For each feature text thus obtained, we put its corresponding label together with the feature text to form a feature-label text. We then put these feature-label texts together into a long text. More details can be found in the Python-like pseudo-code in Listing 1.1 below.



\lstset{style=mystyle}
\label{alg:prompt_gen}
\begin{lstlisting}[language=Python, caption=Generating prompts for GPT4MIA.]
def Prompt_Construct_Part1(F,Y,selection_ratio=0.25): \\ only run once
    ot1=""; m=len(F); k = selection_ratio * m; rep=np.zeros(m,1)
    for i in range(m):
        for j in range(m):
            rep[i]=rep[i]+cosine_sim(f[i],f[j]))
    ind=argsort(rep,"descend"); ind=ind[0:k];    
    for i in reversed(range(k)):
        ot1 = ot+str(f[ind[i]]) + " is in class " 
        + str(argmax(y[ind[i]]) + "\n")
    return ot1
    
def Prompt_Construct_Part2(f_test): \\for each test sample
    ot2= str(f_test) + "is in class \n"
    return ot2          
\end{lstlisting}

\subsection{Workflow and Use Cases}

\begin{figure}[t]
\centering
\includegraphics[width=1.0\textwidth]{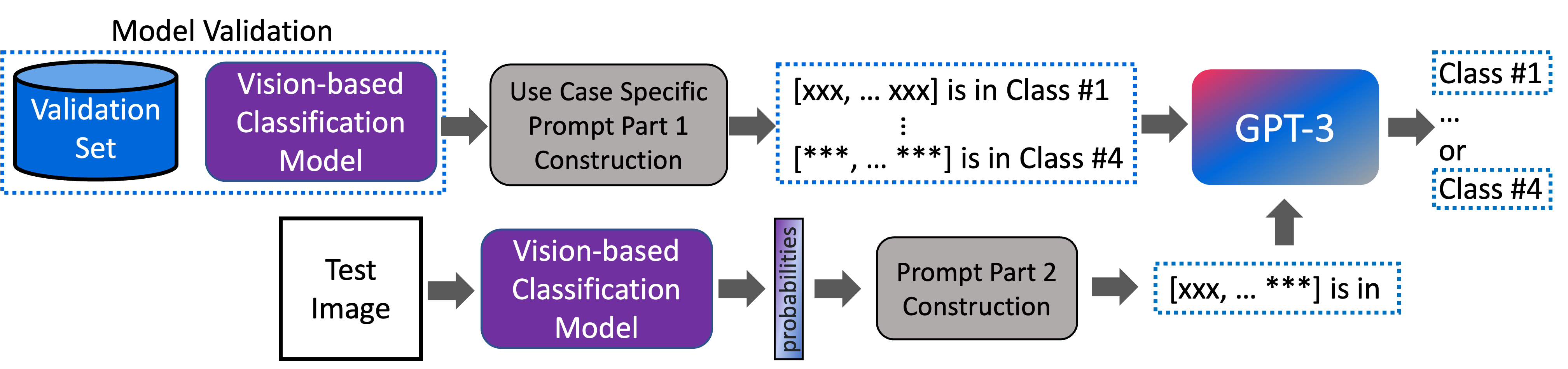}
\caption{The workflow of GPT4MIA. A validation set provides references for transductive inference. }
\label{fig:workflow}
\end{figure}


In this section, we propose two use cases for improving an already-trained vision-based classification model with our proposed GPT4MIA method. The main workflow is illustrated in Fig.~\ref{fig:workflow}.

\textbf{Use Case \#1: Detecting Prediction Errors.} The first use case of utilizing GPT-3 as a transductive inference method is for detecting prediction errors by a trained vision-based classifier. Conventionally, a validation set is commonly used for comparing and selecting models. Here, we 
utilize a validation set to provide known samples for transductive inference. Feature vectors in $F$ are obtained from the output probabilities of the vision-based classification model, and labels in $Y$ are obtained by checking whether the classification model gives the correct prediction on each validation sample. 

\textbf{Use Case \#2: Improving Classification Accuracy.}
The second use case aims to improve an already-trained classifier by directly adjusting its predictions. This is a more challenging scenario in which the method not only seeks to detect wrong predictions but also acts to convert them into correct ones. Feature vectors in $F$ are obtained from the output probabilities of the vision-based classification model, and labels in $Y$ are obtained from the validation set for each validation sample. 






\section{Experiments}
In this section, we empirically validate the effectiveness of our proposed GPT4MIA. Inductive methods (e.g., Linear Regression (LR)~\cite{weisberg2005applied}, Multi-Layer Perception (MLP)~\cite{he2015delving}, and Support Vector Machine (SVM)~\cite{chang2011libsvm}) and transductive methods (e.g., K-Nearest Neighbor (KNN)~\cite{peterson2009k} and Underbagging KNN (UbKNN)~\cite{hang2022under}) are applicable to the two use cases presented above. We compare these methods with GPT4MIA in the experiments below.\footnote{LR, MLP, SVM, and KNN are conducted using the scikit-learn library at https://scikit-learn.org/, and UbKNN is with our implementation.}

\textbf{Configurations:} We use the OpenAI API~\cite{OpenAIAPI} for querying the GPT-3 service for all the experiments related to GPT4MIA. More specifically, the text-Davinci-003 model is used, which can process up to 4000 tokens per request. The hyper-parameter $k$ (for top $k$) is chosen to be a quarter of the number of the total available known samples ($m$). Inference for one test sample costs about \$0.05 USD (charged by OpenAI). For the compared methods, we test their default settings as well as other hyper-parameter settings to report their best results.


\begin{table*}[t!]
\scriptsize
\begin{center}
\caption{Experiments for Use Case \#1: Detecting Prediction Errors. }
\begin{tabular}{||c || c | c | c | c | c | c | c | c ||} 
 \hline
 \multirow{2}{*}{Method}& \multicolumn{4}{|c|}{RetinaMNIST} & \multicolumn{4}{|c|}{FractureMNIST3D}\\
\cline{2-9}
  & Precision & Recall & F-score & Bal-Accu & Precision & Recall & F-score & Bal-Accu\\
\hline
LR  & \textbf{0.617} & 0.631 & 0.624 & 0.486 & 0.492 & \textbf{0.776} & 0.604 & 0.517\\
\hline
MLP & 0.616 & 0.637 & 0.626 & 0.488 & 0.571 & 0.482 & 0.523 & 0.572\\
\hline
SVM  & 0.607 & 0.648 & 0.627 & 0.489 & 0.527 & 0.414 & 0.464 & 0.533\\
\hline

KNN & 0.608 & 0.407 & 0.488 & 0.458 & 0.488 & 0.716 & 0.580 & 0.507\\
\hline

UbKNN & 0.574 & 0.859 & 0.689 & 0.551 & 0.673 & 0.673 & 0.673 & 0.564 \\
\hline
GPT4MIA  & 0.581 & \textbf{0.860} & \textbf{0.693} & \textbf{0.679} & \textbf{0.706} & {0.673} & \textbf{0.689} & \textbf{0.603}\\
\hline

\end{tabular}
\label{table:Task1}

\end{center}
\end{table*}








\begin{table*}[t!]
\scriptsize
\begin{center}
\caption{Experiments for Use Case \#2. Dataset: RetinaMNIST.}
\begin{tabular}{||c || c | c | c | c | c| c||  } 
 \hline
Method  & Class \#1  & Class \#2 & Class \#3 & Class \#4 & Class \#5 & Bal-Accu\\
\hline
N/A  & 0.813 & 0.063 & 0.400 & 0.563 & 0.0 & 0.368\\
\hline
 LR & 0.753 & 0.0 & 0.663 & 0.29 & 0.0 & 0.342\\
  \hline
 MLP & 0.736 & 0.0 & 0.456 & 0.50 & 0.0 & 0.338\\
 \hline
 SVM & 0.729 & 0.0 & 0.369 & 0.75 & 0.0 & 0.370\\
\hline
UbKNN & 0.747 & 0.130 & 0.478 & 0.603 & 0.0 & 0.392\\
\hline 
GPT4MIA  & 0.672 & 0.437 & 0.207 & 0.529 & 0.150 & \textbf{0.399}\\
\hline

\end{tabular}
\label{table:RetinaTask2}
\end{center}
\end{table*}

\begin{table*}[t!]
\scriptsize
\begin{center}
\caption{Experiments for Use Case \#2. Dataset: FractureMNIST3D.}
\begin{tabular}{||c || c | c | c | c ||  } 
 \hline
Method  & Class \#1  & Class \#2 & Class \#3 & Bal-Accu\\
\hline
N/A  & 0.778 & 0.375 & 0.326 & 0.493 \\
\hline
 LR & 0.600 & 0.596 & 0.304 & 0.500\\
  \hline
 MLP & 0.556 & 0.673 & 0.283 & 0.504\\
 \hline
 SVM & 0.533 & 0.596 & 0.391 & 0.507 \\
\hline
UbKNN & 0.644 & 0.394 & 0.478 & 0.505\\
\hline 
GPT4MIA  & 0.522 & 0.510 & 0.543 & \textbf{0.525}\\
\hline

\end{tabular}
\label{table:Frac3DTask2}
\end{center}
\end{table*}

\begin{table*}[t!]
\scriptsize
\begin{center}
\caption{Ablation study for Use Case \#2. Dataset: RetinaMNIST.}
\begin{tabular}{||c || c | c | c | c | c| c||  } 
 \hline
Method  & Class \#1  & Class \#2 & Class \#3 & Class \#4 & Class \#5 & Bal-Accu\\
\hline
w/o Selection \& Ordering & 0.724 & 0.086 & 0.435 & 0.456 & 0.1 & 0.360\\
\hline
GPT4MIA (Full)  & 0.672 & 0.437 & 0.207 & 0.529 & 0.150 & \textbf{0.399}\\
\hline

\end{tabular}
\label{table:AB:RetinaTask2}
\end{center}
\end{table*}

\begin{table*}[t!]
\scriptsize
\begin{center}
\caption{Ablation study for Use Case \#2. Dataset: FractureMNIST3D.}
\begin{tabular}{||c || c | c | c | c ||  } 
 \hline
Method  & Class \#1  & Class \#2 & Class \#3 & Bal-Accu\\
\hline
w/o Selection \& Ordering & 0.311 & 0.769 & 
0.283 & 0.454\\
\hline
GPT4MIA (Full)  & 0.522 & 0.510 & 0.543 & \textbf{0.525}\\
\hline

\end{tabular}
\label{table:AB:Frac3DTask2}
\end{center}
\end{table*}

\subsection{On Detecting Prediction Errors} 
We utilize the RetinaMNIST and FractureMNIST3D datasets from the MedMNIST dataset~\cite{yang2023medmnist} for these experiments. We apply a ResNet-50 model trained with the training set as the vision-based classifier, for which the weights can be obtained from the official release.\footnote{The model weights are obtained from https://github.com/MedMNIST/experiments.} We then collect the model's output probabilities for each validation sample and label it based on whether the prediction is correct. An error detection method is then built based on the information from the validations for classifying the predictions into two classes (being correct or incorrect). The error detection model is then evaluated using the test set working with the same prediction model which was used on the validation (ResNet-50 in this case). We compare our proposed GPT4MIA method on this task with a set of well established inductive methods and transductive methods. From Table~\ref{table:Task1}, one can see that GPT4MIA significantly outperforms the known competing methods for detecting prediction errors from a CNN-based classifier. 

\subsection{On Improving Classification Accuracy}
We utilize the RetinaMNIST and FractureMNIST3D datasets from MedMNIST~\cite{yang2023medmnist} for these experiments. ResNet-50 is used as the trained vision-based classification model. The model weights are obtained from the MedMNIST official release. 
In Table~\ref{table:RetinaTask2}, we observe that GPT4MIA performs similarly when comparing with the state-of-the-art transductive inference method Underbagging KNN in balanced accuracy. In Tabel~\ref{table:Frac3DTask2}, we observe that GPT4MIA performs considerably better in balanced accuracy.

\subsection{Ablation Studies}
We validate the effect of performing sample selection and ordering described in Section~\ref{sec:promptdesign}. In Table~\ref{table:AB:RetinaTask2} and Table~\ref{table:AB:Frac3DTask2}, we show the performances for the setting without the step of sample selection and ordering. From these results, it is clear that sample selection and ordering is important for better performance when utilizing GPT-3 as a transductive inference tool.

\section{Discussion and Conclusions}

In this paper, we developed a novel method called GPT4MIA that utilizes a pre-trained large language model (e.g., GPT-3) for transductive inference for medical image classification. Our theoretical analysis and technical developments are well-founded, and empirical results demonstrated that our proposed GPT4MIA is practical and effective. Large language models (LLMs) such as GPT-3 and, recently, ChatGPT~\cite{chatgptURL} have shown great capability and potential in many different AI applications. In this work, we showed that GPT-3 can perform transductive inference for medical image classification with better accuracy than conventional and state-of-the-art machine learning methods. 
LLMs are great new technologies that can push the boundaries of AI research; on the other hand, new concerns are raised in using these generative models. Reliability and privacy are among the top priorities for medical image analysis, and more efforts should be put into this frontier when working with LLMs. In addition, further improving LLMs for medical image analysis, including better robustness and accuracy, lower costs, and more use cases, are all exciting and important future research targets.
 
\bibliographystyle{plain}
\bibliography{ref}


\newpage
\section{Appendix}
\subsection{Additional Results and Visualizations}
\begin{figure}[]
\centering
\includegraphics[width=0.7\textwidth]{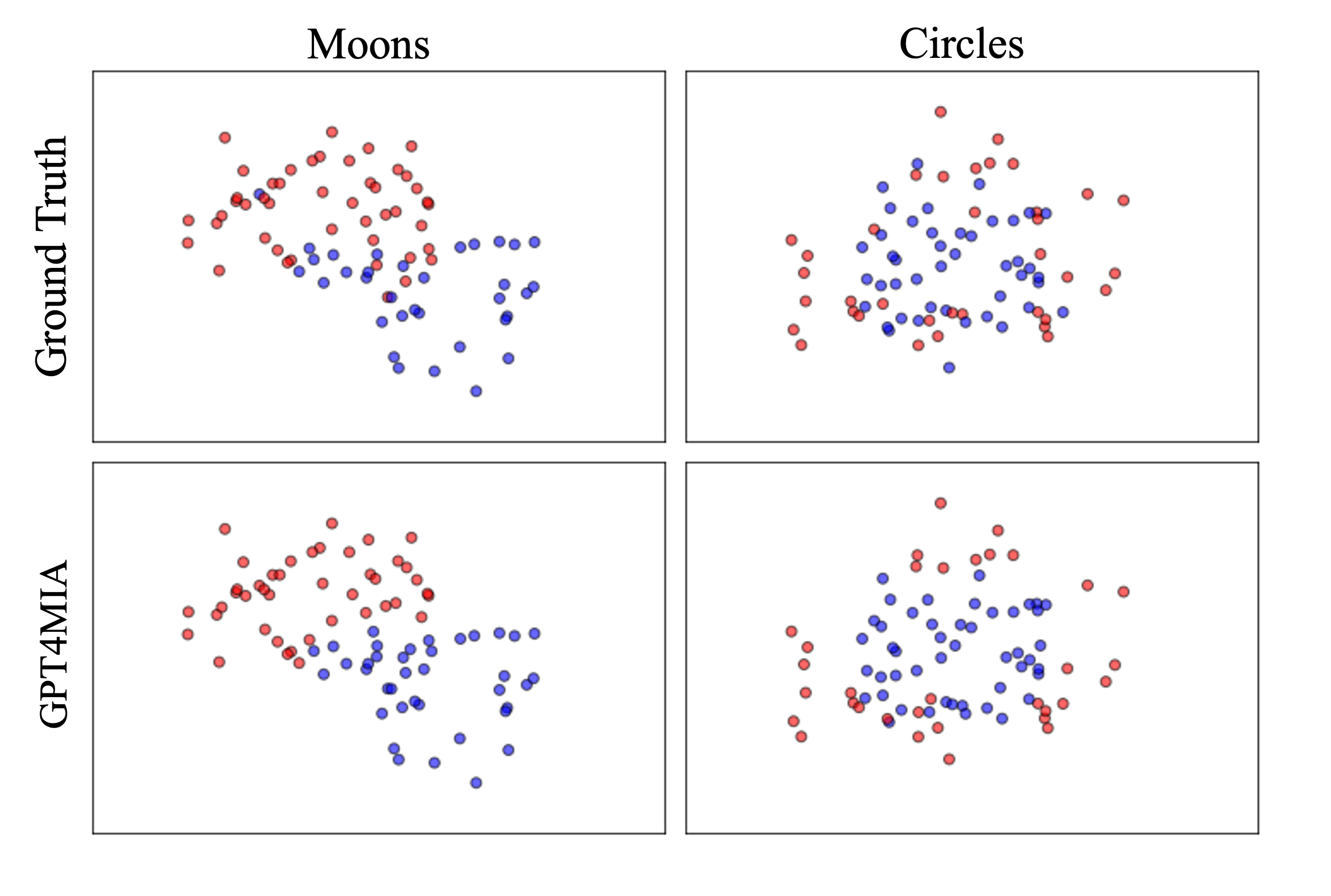}
\caption{Visualizations of classification results (test sample points) on two toy datasets from the scikit-learn package. Different colors indicate different classes.} \label{fig:vis1}
\end{figure}

\begin{figure}[]
\centering
\includegraphics[width=1.0\textwidth]{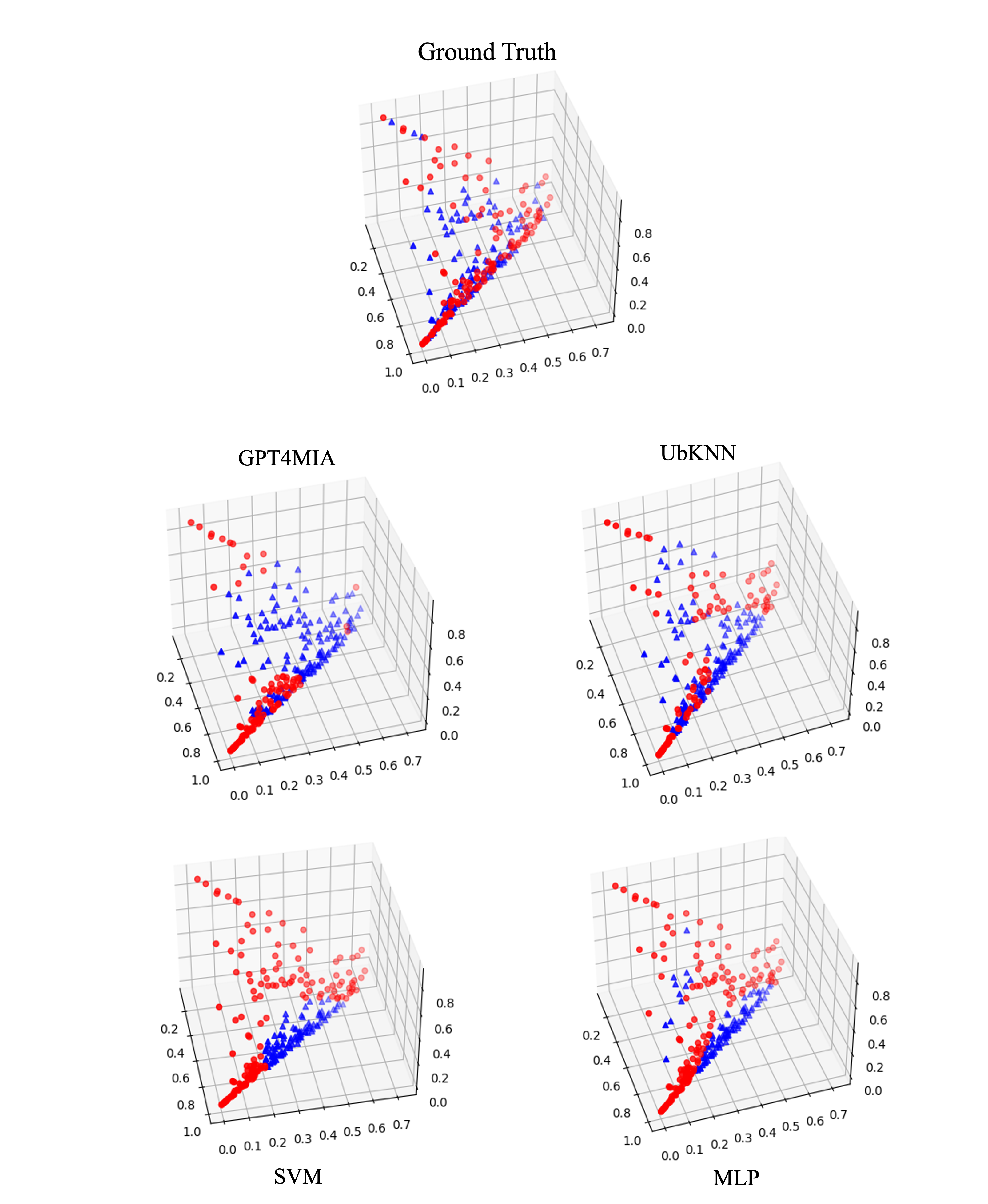}
\caption{Visualizations of test sample points from the experiments for Use Case \#1 (detecting prediction errors) on the FractureMNIST3D dataset. In the ground truth, red points are correct predictions, and blue triangles are incorrect predictions, all by the ResNet-50. In each result sub-figure (e.g., by GPT4MIA), red points are those classified as correct predictions, and blue triangles are those classified as incorrect predictions.} \label{fig:vis2}
\end{figure}

In Table~\ref{table:additional_result1}, we give additional results of GPT4MIA and other well-known classification models on toy datasets from the scikit-learn package. This experiment serves as a sanity check for our proposed GPT4MIA. In Fig.~\ref{fig:vis1}, we visualize the classification results (test sample points) for GPT4MIA and the ground truth. In addition, in Fig.~\ref{fig:vis2}, we give visualizations of the test sample points from the experiments for Use Case \#1 on the FractureMNIST3D dataset. 

\begin{table*}[]
\begin{center}
\scriptsize
\caption{Binary classification results on toy datasets in the scikit-learn package. NearN: Nearest Neighbors; GP: Gaussian Process; DT: Decision Tree; RF: Random Forest; NN: Neural Networks; NB: Naive Bayes; QDA: Quadratic Discriminant Analysis. }
\begin{tabular}{|c | c | c | c | c | c| c| c| c| c| c| c |} 
 \hline
 {Dataset}  & {NearN} & {Linear SVM} & {RBF SVM} & {GP} & {DT} & {RF} & NN & AdaBoost & NB & QDA & GPT4MIA\\
\hline
{Moons}  & 0.86 & 0.85 & 0.89 & 0.86 & 0.83& 0.88 & 0.86 & 0.88 & 0.86 & 0.86 & 0.88\\
\hline
{Circles}  &0.61&0.45&0.68&0.69&0.71&0.71&0.68&0.69&0.63&0.65& {0.81}\\
\hline
\end{tabular}
\label{table:additional_result1}
\end{center}
\end{table*}

\vspace{-1.5cm}

\end{document}